\documentclass[letterpaper,10pt,conference]{ieeeconf}

\IEEEoverridecommandlockouts
\usepackage{amsmath,amssymb,amsfonts}
\usepackage{booktabs}
\usepackage{capt-of}
\usepackage{cite}
\usepackage{graphicx}
\usepackage[table]{xcolor}

\newcommand{\method}{FeasibleFlow}
\newcommand{\sg}{\operatorname{sg}}
\newcommand{\arranker}{Anchor-relative ranker}
\newcommand{\imranker}{Imitation ranker}
\newcommand{\outranker}{Outcome ranker}
\newcommand{\pos}[1]{{\textcolor[rgb]{0.1,0.6,0.1}{\scriptsize\,+#1}}}

\title{\LARGE \bf
\method: One-Step Joint Transport of Configuration Feasibility and Trajectories for End-to-End Driving
}

\author{
Xiang Li\textsuperscript{\rm 1},
Bikun Wang\textsuperscript{\rm 1}\dag\thanks{\dag~Bikun Wang is the Corresponding author.},
Qing Xu\textsuperscript{\rm 1},
Jianjun Wang\textsuperscript{\rm 1} \\
\textsuperscript{1} Cross-Domain Computing Solutions, Bosch
}

\begin{document}
\maketitle
\thispagestyle{empty}
\pagestyle{empty}

\begin{abstract}
End-to-end autonomous driving maps current observations directly to future trajectories, yet those trajectories must remain valid as the scene evolves. Future state modeling aims to address this temporal mismatch, but general representations often contain information unrelated to ego planning and affect trajectory generation only through auxiliary supervision, static conditioning, or proposal evaluation. We propose \method, a one-step end-to-end generative framework that jointly transports a configuration-space feasibility field and multimodal ego trajectories. Our Asymmetric Joint MeanFlow uses the pathwise Jacobian-vector product in the MeanFlow identity to incorporate field evolution into trajectory transport. Because safety feedback is sparser than progress feedback, we further introduce the \arranker{} (ARR) and Pareto-ReinFlow to balance safety and progress in candidate selection and generation, respectively. Experiments on the NAVSIM benchmark demonstrate the strong performance of \method{} and validate both the joint transport of feasibility and trajectories and the proposed safety-first mechanisms.
\end{abstract}

\section{INTRODUCTION}

\begin{figure}[t]
    \centering
    \includegraphics[width=\columnwidth]{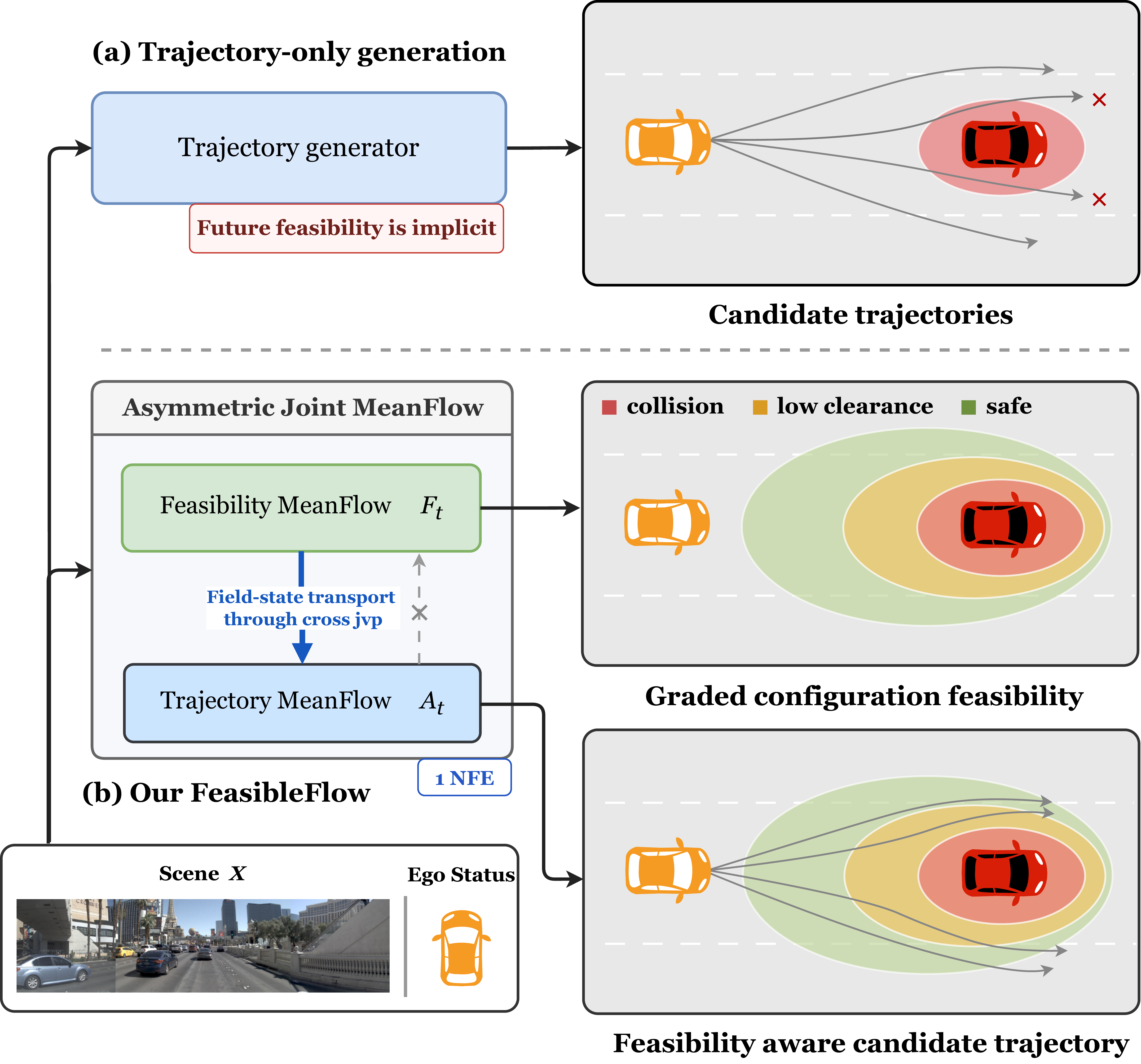}
    \caption{\textbf{Motivation and key mechanism of \method.} A trajectory-only generator leaves future feasibility implicit. \method{} instead jointly transports a graded configuration-space feasibility field and multimodal ego actions. The evolving field enters action transport through the cross-JVP term of the Asymmetric Joint MeanFlow, while inference remains a single transport step.}
    \label{fig:motivation}
\end{figure}

End-to-end (E2E) driving predicts future motion from current observations, yet the plan must remain safe as the scene evolves. Structured BEV, occupancy, and vectorized representations expose scene information to the planner \cite{hu2022stp3,hu2023uniad,jiang2023vad,wang2023interpretable}. Future-aware methods generate scene states, roll out action-conditioned worlds, or estimate proposal outcomes \cite{zheng2024genad,yang2025driveoccworld,li2025wote,zheng2025world4drive,zhang2025seerdrive,fu2026prodrive}, while generative planners model multimodal ego motion \cite{liao2025diffusiondrive,xing2025goalflow,liu2025guideflow,wang2026meanfuser,xu2026wamflow}. However, future representations usually enter trajectory generation through auxiliary objectives, static conditioning, or proposal evaluation. Their broad scene descriptions leave the geometric constraints imposed by the ego footprint and orientation implicit. Figure~\ref{fig:motivation} illustrates this gap.

We formulate future geometric constraints in the ego vehicle's configuration space. A candidate pose is feasible only if the oriented ego footprint can be placed there with sufficient clearance from surrounding constraints. Configuration-space geometry projects drivable boundaries, static obstacles, and future traffic participants into this planning-oriented form \cite{lozanoperez1983,ratliff2009chomp}. We encode these constraints as a graded feasibility field that supports sparse, trajectory-relevant queries rather than dense future-world reconstruction.

We therefore propose \method, an end-to-end framework that jointly transports the feasibility field and multimodal ego trajectories in one step. MeanFlow learns interval-average velocity and relates it to the underlying dynamics through the pathwise directional derivative in its identity \cite{geng2025meanflow}. By treating field and action variables as a joint state, the Asymmetric Joint MeanFlow makes the action-training target depend on synchronous field evolution while keeping field transport action-independent. An implicit codec learns the field state from sparse configuration queries, and an exponential moving average (EMA) encoder provides a stable transport target. Future geometric constraints thus enter trajectory generation without sacrificing one-step inference.

Multimodal generation still requires a final choice between safety and progress. Because progress feedback is denser than collision and drivable-area feedback, a learned ranker may overfavor high-progress candidates. The \arranker{} (ARR) uses the \imranker{} to establish an imitation anchor, then uses the \outranker{} to seek progress without predicted safety degradation relative to that anchor. Since ranking cannot change the candidate distribution, Pareto-ReinFlow further updates action transport with exact, safety-first planning feedback.

Our contributions are summarized as follows:
\begin{itemize}
    \item We represent clearance conditioned on the ego footprint and orientation as a compact configuration-space feasibility field, learned from sparse graded queries by an implicit codec.
    \item We introduce an Asymmetric Joint MeanFlow whose block-triangular dependency and joint Jacobian-vector product (JVP) inject field evolution into trajectory learning while retaining one-step inference.
    \item We combine the \arranker{} with Pareto-ReinFlow to improve candidate selection and the generation distribution under safety-first relative feedback.
\end{itemize}

\begin{figure*}[t]
    \centering
    \includegraphics[width=\textwidth]{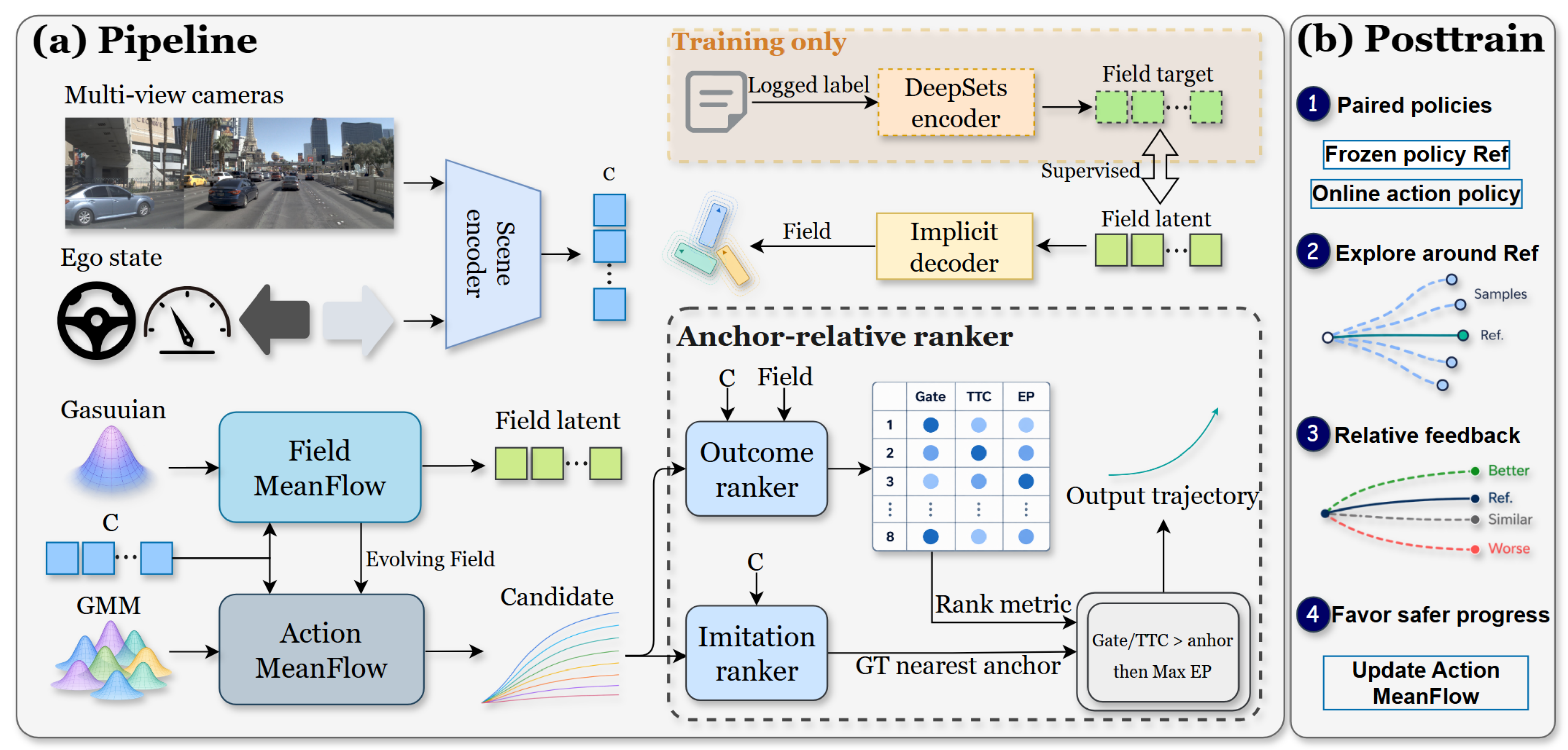}
    \caption{\textbf{Overall architecture and training pipeline of \method.} (a) During supervised training, a DeepSets encoder and implicit decoder learn the feasibility field from logged configuration labels. Field and action states are transported by the Asymmetric Joint MeanFlow, and the \arranker{} (ARR) selects the final candidate through its \imranker{} and \outranker. (b) Pareto-ReinFlow uses the frozen ranker to select a reference trajectory, evaluates exploratory trajectories by their relative safety and progress, and updates only the online action MeanFlow.}
    \label{fig:architecture}
\end{figure*}

\section{RELATED WORK}

\subsection{End-to-End Driving and Future-World Modeling}
End-to-end driving frameworks differ mainly in scene representation and task organization. TransFuser fuses camera and LiDAR features \cite{chitta2022transfuser}, while ST-P3 connects perception, prediction, and planning through interpretable maps, occupancy, and spatiotemporal BEV features \cite{hu2022stp3}. UniAD coordinates driving tasks with planning-oriented queries, whereas VAD uses vectorized agents and map elements as explicit planning constraints \cite{hu2023uniad,jiang2023vad}. PARA-Drive parallelizes mapping, prediction, and planning, and SparseDrive unifies the pipeline with sparse instance representations \cite{weng2024paradrive,sun2024sparsedrive}.

Future-world methods differ in what future state the planner consumes. GenAD jointly models ego and agent motion in a structured trajectory latent space, whereas OccWorld jointly predicts ego motion and three-dimensional semantic occupancy \cite{zheng2024genad,zheng2024occworld}. Drive-OccWorld evaluates actions through predicted occupancy and scene flow \cite{yang2025driveoccworld}; World4Drive predicts intent-conditioned latent states, and WoTE estimates candidate execution outcomes \cite{zheng2025world4drive,li2025wote}. These methods expose future agent trajectories, occupancy, BEV features, or latent states to planning. \method{} instead represents clearance conditioned on the ego footprint and orientation, jointly transports this configuration-space field with ego actions, and uses it in candidate evaluation.

\subsection{Generative Trajectory Planning and Policy Post-training}
Complex traffic scenes admit multiple reasonable actions, motivating generative trajectory models based on diffusion or flow. Diffusion-ES combines a diffusion prior with gradient-free search, DiffusionDrive truncates denoising around trajectory anchors, and Diffusion Planner jointly generates ego and agent motion \cite{yang2024diffusiones,liao2025diffusiondrive,zheng2025diffusionplanner}. Flow matching instead models generation as continuous transport \cite{lipman2023flow}: GoalFlow adds goal guidance, GuideFlow introduces physical and energy-based constraints, WAM-Flow transports discrete trajectory tokens, and MeanFuser uses MeanFlow for one-step multimodal planning \cite{xing2025goalflow,liu2025guideflow,xu2026wamflow,wang2026meanfuser}.

Policy post-training adjusts a learned generator with external feedback. ReinFlow introduces learnable noise to obtain tractable transition probabilities for few-step or one-step flow policies, whereas DriveDPO optimizes trajectory preferences derived from driving rules \cite{zhang2025reinflow,shang2025drivedpo}. \method{} retains one-step MeanFlow generation but conditions action transport on configuration-space feasibility. Pareto-ReinFlow then uses the output of the frozen \arranker{} as a reference and updates the candidate distribution through relative safety ordering.

\section{PRELIMINARIES}

\subsection{Problem Formulation}
End-to-end autonomous driving maps multi-view sensor observations, route information, and ego state $X$ directly to a future trajectory:
\begin{equation}
Y=(\mathbf y_1,\ldots,\mathbf y_T), \qquad
\mathbf y_\tau=(p_\tau^x,p_\tau^y,\psi_\tau),
\label{eq:trajectory}
\end{equation}
where $\mathbf y_\tau$ denotes the planar ego pose, including two-dimensional position and heading, at future time step $\tau$.

Conventional generative planners typically model the conditional trajectory distribution $p(Y\mid X)$. \method{} instead jointly models a future configuration-space feasibility field and the corresponding ego response. We introduce a scene-level future feasibility state $F\in\mathbb R^{T\times D}$ and factorize the joint distribution as:
\begin{equation}
p(F,Y\mid X,m)=p(F\mid X)\,p(Y\mid F,X,m),
\label{eq:factorization}
\end{equation}
where $m$ denotes the behavior hypothesis associated with a multimodal action source. This factorization separates scene-level future feasibility from the mode-dependent ego response.

\subsection{MeanFlow}
Given a target state $z_0$ and a source state $z_1$, we define the linear path and its direction as:
\begin{equation}
z_t=(1-t)z_0+t z_1, \qquad V=z_1-z_0.
\label{eq:linear_path}
\end{equation}
Let $v(z_\xi,\xi)$ denote the instantaneous velocity along the path. MeanFlow parameterizes the average velocity over an interval $[r,t]$, rather than directly parameterizing the instantaneous velocity:
\begin{equation}
u(z_t,r,t)=\frac{1}{t-r}\int_r^t v(z_\xi,\xi)\,\mathrm d\xi,
\qquad 0\le r<t\le1.
\label{eq:average_velocity}
\end{equation}
Differentiating $(t-r)u(z_t,r,t)$ with respect to $t$ gives the MeanFlow identity:
\begin{equation}
u(z_t,r,t)+(t-r)D u(z_t,r,t)[V,0,1]=v(z_t,t),
\label{eq:meanflow_identity}
\end{equation}
where $D u[V,0,1]$ is the JVP computed with $r$ fixed along the direction $(V,0,1)$. The JVP measures the instantaneous change in the network output as its input moves along the specified path. Since the linear path satisfies $v(z_t,t)=V$, the stop-gradient regression target is:
\begin{equation}
\widehat u=\sg\!\left(V-(t-r)D u_\theta(z_t,r,t)[V,0,1]\right).
\label{eq:mf_target}
\end{equation}
The model is optimized by minimizing:
\begin{equation}
\mathcal L_{\mathrm{MF}}=\mathbb E\!\left[\lVert u_\theta-\widehat u\rVert_2^2\right].
\label{eq:mf_loss}
\end{equation}
At inference, setting $(r,t)=(0,1)$ gives the one-step estimate:
\begin{equation}
\widehat z_0=z_1-u_\theta(z_1,0,1).
\label{eq:mf_inference}
\end{equation}

\section{APPROACH}
\label{sec:approach}

\subsection{Overview}
\method{} is an end-to-end framework whose training pipeline comprises representation learning, joint generation, candidate selection with the \arranker, and Pareto-ReinFlow policy post-training. Representation learning converts future geometric information from the log into time-indexed configuration queries and encodes them as the field endpoint $F_0$. The joint generator then uses $F_0$ and the expert action $A_0$ as targets to learn an Asymmetric Joint MeanFlow on a shared clock, allowing the action path to read the evolving field state. The \arranker{} uses the \imranker{} to establish an imitation anchor and the \outranker{} to select a higher-progress trajectory without predicted safety degradation relative to that anchor. After supervised training, the perception and field modules are frozen, and Pareto-ReinFlow updates only the action transport using exact planning feedback. Figure~\ref{fig:architecture} summarizes the roles of these stages during training and inference.

\newcommand{\navsimresulttables}{%
\begin{table*}[t]
\centering
\caption{
\textbf{Planning performance on NAVSIM v1 \texttt{navtest}.}
}
\label{tab:navsim_v1}
\setlength{\tabcolsep}{8pt}
\renewcommand{\arraystretch}{1.15}

\begin{small}
    \begin{tabular}{l cc ccccc >{\columncolor[HTML]{E6E6E6}}c}
        \toprule
        \textbf{Method}
        & \textbf{Input}
        & \textbf{Backbone}
        & \textbf{NC} $\uparrow$
        & \textbf{DAC} $\uparrow$
        & \textbf{TTC} $\uparrow$
        & \textbf{Comf.} $\uparrow$
        & \textbf{EP} $\uparrow$
        & \textbf{PDMS} $\uparrow$ \\
        \midrule
        
        TransFuser~\cite{chitta2022transfuser}
        & C \& L
        & ResNet-34
        & 97.7 & 92.8 & 92.8 & \textbf{100.0} & 79.2 & 84.0 \\

        Hydra-MDP~\cite{li2024hydramdp}
        & C \& L
        & ResNet-34
        & 98.3 & 96.0 & 94.6 & \textbf{100.0} & 78.7 & 86.5 \\
        
        GoalFlow~\cite{xing2025goalflow}
        & C \& L
        & ResNet-34
        & 98.3 & 93.8 & 94.3 & \textbf{100.0} & 79.8 & 85.7 \\

        DiffusionDrive~\cite{liao2025diffusiondrive}
        & C \& L
        & ResNet-34
        & 98.2 & 96.2 & 94.7 & \textbf{100.0} & 82.2 & 88.1 \\

        WoTE~\cite{li2025wote}
        & C \& L
        & ResNet-34
        & \textbf{98.5} & 96.8 & \textbf{94.9} & 99.9 & 81.9 & 88.3 \\

        SeerDrive~\cite{zhang2025seerdrive}
        & C \& L
        & ResNet-34
        & 98.4 & 97.0 & \textbf{94.9} & 99.9 & 83.2 & 88.9 \\

        UniAD~\cite{hu2023uniad}
        & C
        & ResNet-34
        & 97.8 & 91.9 & 92.9 & \textbf{100.0} & 78.8 & 83.4 \\

        World4Drive~\cite{zheng2025world4drive}
        & C
        & ResNet-34
        & 97.4 & 94.3 & 92.8 & \textbf{100.0} & 79.9 & 85.1 \\

        Epona~\cite{zhang2025epona}
        & C
        & Transformer
        & 97.9 & 95.1 & 93.8 & 99.9 & 80.4 & 86.2 \\
        
        MeanFuser~\cite{wang2026meanfuser}
        & C
        & ResNet-34
        & 98.2 & 97.5 & 94.4 & \textbf{100.0} & 83.3 & 89.0 \\
        
        \midrule
        
        \method{} (\textbf{Ours})
        & C
        & ResNet-34
        & 97.9 & \textbf{98.6} & 93.8 & 99.8 & \textbf{85.4} & \textbf{90.1} \\
        
        \bottomrule
    \end{tabular}
\end{small}
\vspace{-0.1in}
\end{table*}

\begin{table*}[t]
    \centering
    \setlength{\tabcolsep}{6pt}
    \caption{
        \textbf{Planning performance on NAVSIM v2 \texttt{navtest}.}
    }
    \label{tab:navsim_v2}
    \vspace{2pt}

    \begin{small}
        \renewcommand{\arraystretch}{1.15}
        \begin{tabular}{l ccccccccc >{\columncolor[HTML]{E6E6E6}}c}
            \toprule
            \textbf{Method}
            & \textbf{NC} $\uparrow$
            & \textbf{DAC} $\uparrow$
            & \textbf{DDC} $\uparrow$
            & \textbf{TLC} $\uparrow$
            & \textbf{EP} $\uparrow$
            & \textbf{TTC} $\uparrow$
            & \textbf{LK} $\uparrow$
            & \textbf{HC} $\uparrow$
            & \textbf{EC} $\uparrow$
            & \textbf{EPDMS} $\uparrow$ \\
            \midrule

            TransFuser~\cite{chitta2022transfuser}
            & 96.9
            & 89.9
            & 97.8
            & \underline{99.7}
            & 87.1
            & 95.4
            & 92.7
            & \textbf{98.3}
            & 87.2
            & 76.7 \\

            Hydra-MDP++~\cite{li2025hydramdpp}
            & 97.2
            & 97.5
            & \underline{99.4}
            & 99.6
            & 83.1
            & 96.5
            & 94.4
            & \underline{98.2}
            & 70.9
            & 81.4 \\

            DriveSuprim~\cite{yao2026drivesuprim}
            & 97.5
            & 96.5
            & \underline{99.4}
            & 99.6
            & \underline{88.4}
            & 96.6
            & 95.5
            & \textbf{98.3}
            & 77.0
            & 83.1 \\

            DiffusionDriveV2~\cite{zou2025diffusiondrivev2}
            & 97.7
            & 96.6
            & 99.2
            & \textbf{99.8}
            & \textbf{88.9}
            & 97.2
            & 96.0
            & 97.8
            & \textbf{91.0}
            & 85.5 \\

            Drive-JEPA~\cite{wang2026drivejepa}
            & \textbf{98.4}
            & \underline{98.6}
            & 99.1
            & \textbf{99.8}
            & \underline{88.4}
            & \textbf{97.8}
            & \textbf{97.6}
            & 97.9
            & 84.8
            & 87.8 \\

            MeanFuser~\cite{wang2026meanfuser}
            & \underline{98.3}
            & 97.2
            & \textbf{99.6}
            & \textbf{99.8}
            & 87.6
            & \underline{97.4}
            & \underline{97.3}
            & \textbf{98.3}
            & \underline{88.2}
            & \underline{89.5} \\
            \midrule

            \method{} (\textbf{Ours})
            & 97.8
            & \textbf{98.7}
            & 99.2
            & \underline{99.7}
            & \textbf{88.9}
            & 97.1
            & 96.1
            & 97.8
            & 85.1
            & \textbf{89.7} \\
            \bottomrule
        \end{tabular}
    \end{small}
\end{table*}
}

\subsection{Implicit Configuration Feasibility}
We define the geometric constraints of the logged future in the ego vehicle's configuration space and construct feasibility queries at each future time step. For a query pose $q=(x,y,\psi)$ at time $\tau$, let the ego footprint be $P(q)$. Inflating the footprint by the Minkowski sum with a disk of radius $\mu$ gives $P_\mu(q)=P(q)\oplus B(\mu)$. We define a binary query label $b_\tau(q,\mu)$ that indicates whether the ego footprint, with its actual size and orientation, lies within the drivable region without geometrically intersecting static obstacles or future dynamic traffic participants. We use four inflation radii, $\mathcal M=\{0,0.5,1.0,2.0\}\,\mathrm m$, and combine the binary labels into a five-level ordinal variable:
\begin{equation}
c_\tau(q)=\sum_{\mu\in\mathcal M}b_\tau(q,\mu)\in\{0,1,2,3,4\}.
\label{eq:ordinal_feasibility}
\end{equation}
Under this definition, class 0 denotes a collision or an invalid position, and a larger class value indicates greater safety clearance.

Sparse queries are constructed around the action distribution of the generator. The means of the Gaussian mixture modes form behavior skeletons, and two additional support trajectories are sampled from each mode to cover within-mode variation. At each of the eight future steps of every trajectory, we construct nine queries at the original configuration and at local longitudinal, lateral, and heading perturbations. The lateral offsets are $\{\pm0.5,\pm1.0\}\,\mathrm m$, the longitudinal offsets are $\{\pm0.5\}\,\mathrm m$, and the heading offsets are $\{\pm5^\circ\}$. Queries around the expert trajectory contribute to decoder supervision but are excluded from the set aggregated by the field encoder, reducing the possibility that the field latent directly copies the expert action.

At each time step, the query encoder processes the normalized pose feature $\phi(q)=[x/32,\;y/32,\;\sin\psi,\;\cos\psi]$ together with its ordinal label and applies average pooling to obtain $f_\tau^0\in\mathbb R^{128}$. The eight temporal slices form the clean field endpoint $F_0=\{f_\tau^0\}_{\tau=1}^{8}$. A conditional implicit decoder takes $f_\tau^0$ and the query pose as input and predicts the class probability $p(c\mid f_\tau^0,q)$. Let $\mathcal Q_v$ denote the set of valid queries. The query reconstruction loss is:
\begin{equation}
\begin{aligned}
\mathcal L_{\mathrm{codec}}
&=-\frac{1}{|\mathcal Q_v|\log 5}
\sum_{(\tau,q)\in\mathcal Q_v}\log p\!\left(c_\tau(q)\mid f_\tau^0,q\right).
\end{aligned}
\label{eq:codec_loss}
\end{equation}
Division by $\log 5$ only normalizes the loss scale. Class-tail probabilities recover feasibility probabilities at each inflation level and ensure that feasibility does not increase with the inflation radius. The online encoder learns the field representation through this reconstruction objective, while its EMA copy provides a stop-gradient field endpoint. The query constructor and field encoder are used only during training.

\subsection{Asymmetric Joint MeanFlow}
Predicting the feasibility field independently does not ensure that it participates in trajectory generation. Let $F_0$ and $A_0$ denote the field target and expert action, and let $F_1$ and $A_1^{(m)}$ denote the Gaussian field source and the $m$-th mixture action source, respectively. The field and action share the same transport time:
\begin{equation}
\begin{aligned}
(F_t,A_t)&=(1-t)(F_0,A_0)+t(F_1,A_1^{(m)}),
\end{aligned}
\label{eq:joint_path}
\end{equation}
Their conditional dependencies are asymmetric. The field velocity $u_F(F_t,r,t;X)$ depends only on the evolving field and scene observation, whereas the action velocity $u_A(A_t,h_t,r,t;X,m)$ also depends on the field condition $h_t=h(F_t)$ and trajectory mode $m$. This directionality follows from the supervision structure: a log provides multi-configuration geometric labels for one global field, but does not record how other traffic participants would respond to counterfactual ego actions.

The JVP measures the instantaneous change in the network output as its input moves along a given path. Consider advancing the shared path by an infinitesimal step $\delta$: the field state becomes $F_t+\delta V_F$, the action state becomes $A_t+\delta V_A$, and the transport time becomes $t+\delta$, while the interval start $r$ remains fixed. The JVPs of the two branches are therefore:
\begin{equation}
\mathcal J_F
=\left.\frac{\mathrm d}{\mathrm d\delta}
u_F(F_t+\delta V_F,r,t+\delta;X)\right|_{\delta=0},
\label{eq:field_jvp}
\end{equation}
\begin{equation}
\begin{aligned}
\mathcal J_A
&=\frac{\mathrm d}{\mathrm d\delta}
u_A\!\Bigl(A_t+\delta V_A,h(F_t+\delta V_F),r,t+\delta;X,m\Bigr)\Big|_{\delta=0},
\end{aligned}
\label{eq:action_jvp}
\end{equation}
The first expression advances the field state and transport time together, while the second also advances the action state. Consequently, $\mathcal J_A$ includes the effect of field evolution on action generation. In practice, each JVP is computed in one automatic-differentiation operation and is used only during training.

Let $y_s$ denote the stop-gradient MeanFlow target in Eq.~\eqref{eq:mf_target} for branch $s$. Both branches are optimized with the following $L_1$ loss:
\begin{equation}
\mathcal L_{s\text{-}\mathrm{MF}}=\lVert u_s-y_s\rVert_1,
\qquad s\in\{F,A\}.
\label{eq:branch_mf_loss}
\end{equation}
This coupling makes configuration feasibility part of the action-transport dynamics rather than a static feature outside the generator.

\subsection{Supervised Objective and \arranker}
Supervised training jointly learns generation, field representation, and candidate selection. The action and field MeanFlow losses train the two transport branches, the query reconstruction and BEV semantic losses constrain the field representation, and the ranking loss trains the \arranker. The overall objective is
\begin{equation}
\begin{aligned}
\mathcal L_{\mathrm{sup}}
&=\lambda_A\mathcal L_{A\text{-}\mathrm{MF}}
+\lambda_F\mathcal L_{F\text{-}\mathrm{MF}}\\
&\quad+\lambda_C\mathcal L_{\mathrm{codec}}
+\lambda_S\mathcal L_{\mathrm{sem}}
+\lambda_R\mathcal L_{\mathrm{rank}}.
\end{aligned}
\label{eq:supervised_objective}
\end{equation}
Here, $\lambda_A,\lambda_F,\lambda_C,\lambda_S,$ and $\lambda_R$ are loss weights, $\mathcal L_{\mathrm{codec}}$ is defined in Eq.~\eqref{eq:codec_loss}, and $\mathcal L_{\mathrm{sem}}$ is the BEV semantic loss.

In exact planning evaluation, progress scores vary continuously, whereas collision and drivable-area signals are sparse. Preliminary experiments show that a learned ranker fits progress differences more readily and can favor high-progress candidates at the expense of safety. We therefore introduce the \arranker, which uses the candidate closest to the expert trajectory as an imitation anchor and constrains progress improvement with the expert behavior prior.

The \arranker{} contains two independent branches: the \imranker{} and the \outranker. The \imranker{} identifies the candidate with the lowest average displacement error in each scene and is trained with focal binary cross-entropy $\mathcal L_{\mathrm{anchor}}$. An EMA copy provides the imitation anchor at inference.

The \outranker{} learns the safety and progress ordering between candidates. The exact evaluator provides four components: no-at-fault collision (NC), drivable-area compliance (DAC), time to collision (TTC), and ego progress (EP). The first two are multiplied to form the base safety gate $g=\mathrm{NC}\times\mathrm{DAC}$. For each of $g$, TTC, and EP, all candidate pairs are compared. If candidate $i$ has a higher exact evaluation than candidate $j$, training encourages the predicted ranking values to satisfy $\gamma_i^e>\gamma_j^e$, and vice versa. Let $s_i^e$ denote the exact evaluation value, $e\in\{g,\mathrm{TTC},\mathrm{EP}\}$. The component loss is:
\begin{equation}
\begin{aligned}
    \mathcal L_e ={}& \frac{1}{\sum_{i<j}w_{ij}^e} \sum_{i<j}w_{ij}^e\log\!\Bigg(1 \,+ \\
    &\exp\!\Big[-\operatorname{sign}(s_i^e-s_j^e)(\gamma_i^e-\gamma_j^e)\Big]\Bigg),
\end{aligned}
\label{eq:component_rank_loss}
\end{equation}
The base safety loss $\mathcal L_g$ weights each pair by the difference in $g$. The TTC and EP losses weight each pair by the corresponding component difference multiplied by the smaller $g$ of the two candidates. Candidate pairs with a clearer ordering therefore have greater influence, while pairs without shared safety support provide no TTC or progress supervision. If no valid pair exists, the corresponding loss is set to zero. The complete selection loss is:
\begin{equation}
\begin{aligned}
\mathcal L_{\mathrm{rank}}
&=\frac{1}{2}\Bigl[\mathcal L_{\mathrm{anchor}}+\frac{1}{3}\bigl(
\mathcal L_g+\mathcal L_{\mathrm{TTC}}+\mathcal L_{\mathrm{EP}}
\bigr)\Bigr].
\end{aligned}
\label{eq:rank_loss}
\end{equation}

At inference, the EMA copy of the \imranker{} first selects imitation anchor $a$. The \arranker{} then retains candidates whose predicted $g$ and TTC from the \outranker{} are both no lower than those of the anchor and selects the candidate with the highest predicted EP. The anchor always satisfies the filtering conditions; if multiple candidates have the same EP, the \imranker{} score determines the result. The \outranker{} therefore searches for progress improvement only under a relative safety constraint.

\subsection{Pareto-ReinFlow Policy Post-training}
Demonstration supervision mainly constrains the generation region near the expert trajectory and cannot determine whether other candidates are safe. The \arranker{} can only reorder existing candidates and cannot change the candidate distribution. We therefore update the action MeanFlow with exact planning feedback after supervised training while retaining the original motion modes, field condition, and one-step generation structure. We call this stage Pareto-ReinFlow.

Deterministic MeanFlow does not explicitly provide action probabilities. To compute the probability ratio required for policy updates, we treat each action endpoint as the mean of a diagonal-Gaussian proxy policy and add a bounded exploration scale to the normalized $xy$ action increments \cite{zhang2025reinflow}. The action head and exploration scale from the source checkpoint form a fixed behavior policy $\pi_b$. The online policy $\pi_\theta$ is initialized from the same checkpoint and then updated. For each of the $K=8$ motion modes, we independently sample two action sources. The behavior policy outputs a mean $\mu_i^b$ and generates a training action:
\begin{equation}
\widetilde A_i^{xy}=\mu_i^{b,xy}+\sigma_i^b\odot\epsilon_i,
\qquad \epsilon_i\sim\mathcal N(0,I).
\label{eq:exploration_action}
\end{equation}
The online policy produces $\mu_i^\theta$ from the same action source and field state. Fixing $\pi_b$ keeps both the sampling distribution and the denominator of the probability ratio unchanged throughout post-training, and exploration noise is used only during training.

Training feedback is constructed around one reference trajectory. The frozen \arranker{} first chooses the reference from the first clean set of $K$ candidates. The exact evaluator then evaluates the $2K$ exploratory trajectories and the reference. In addition to NC, DAC, and TTC, the safety evaluation includes collision survival (CIS) and TTC-violation survival (TIS). We collect these five components as:
\begin{equation}
S_i=(\mathrm{NC}_i,\mathrm{DAC}_i,\mathrm{TTC}_i,\mathrm{CIS}_i,\mathrm{TIS}_i).
\label{eq:safety_vector}
\end{equation}
If a candidate is no worse than the reference on every safety component, it enters the improvement tier and is ranked by EP within that tier. The remaining candidates are compared using a strict safety Pareto relation: one candidate wins only if it is no worse on every safety component and better on at least one. EP is used only when all safety components are equal; all other incomparable relations are treated as ties. Progress gains therefore cannot compensate for safety degradation.

We encode each pairwise win, loss, or tie as $1$, $-1$, or $0$, respectively, and denote it by $R_{ij}$. The average comparison outcome of each candidate against the remaining trajectories is then centered by subtracting the corresponding score of the reference trajectory:
\begin{equation}
\begin{aligned}
\beta_i&=\frac{1}{2K}\sum_{j\ne i}R_{ij},\\[-1pt]
\alpha_i&=\operatorname{clip}\!\left(
\frac{\beta_i-\beta_{\mathrm{ref}}}{2},-1,1\right).
\end{aligned}
\label{eq:reference_advantage}
\end{equation}
For the same training action, we compute its log-probabilities $\ell_i^b$ and $\ell_i^\theta$ under the behavior and online policies, respectively, summing over all $8\times2$ coordinates. Their ratio,
\begin{equation}
\rho_i=\exp(\ell_i^\theta-\ell_i^b),
\label{eq:probability_ratio}
\end{equation}
measures the online policy's relative preference for the action: a positive advantage increases its probability, whereas a negative advantage decreases it. To limit the magnitude of each update, we use the clipped objective with $\varepsilon=0.2$:
\begin{equation}
\begin{aligned}
\mathcal L_{\mathrm{post}}
&=-\mathbb E_{i:\alpha_i\ne0}\!\Bigl[
\min\!\Bigl(\rho_i\alpha_i,\operatorname{clip}\!\left(
\rho_i,1-\varepsilon,1+\varepsilon\right)
\alpha_i\Bigr)\Bigr].
\end{aligned}
\label{eq:posttraining_loss}
\end{equation}
The advantages are detached before optimization, and the loss is set to zero when no nonzero advantage is available. Post-training updates only the action MeanFlow and the online exploration scale. At inference, the behavior policy, exploration noise, reference trajectory, and exact evaluator are not used; the framework still performs one clean MeanFlow rollout with $K=8$ candidates followed by candidate selection.

\newcommand{\componentablationtable}{%
\centering
\captionof{table}{
\textbf{Ablation of key components on NAVSIM v1 \texttt{navtest}.}
}
\label{tab:feasibleflow_ablation}
\setlength{\tabcolsep}{3.5pt}
\renewcommand{\arraystretch}{1.15}

\resizebox{\linewidth}{!}{
    \begin{tabular}{c ccc cccc >{\columncolor[HTML]{E6E6E6}}c}
    \toprule
    \textbf{ID}
    & \textbf{Joint}
    & \textbf{ARR}
    & \textbf{PT}
    & \textbf{NC}
    & \textbf{DAC}
    & \textbf{TTC}
    & \textbf{EP}
    & \textbf{PDMS} \\
    \midrule

    $\mathcal{M}_0$
    & $\times$ & $\times$ & $\times$
    & 97.9 & 96.6 & 93.5 & 82.5 & 87.8 \\

    $\mathcal{M}_1$
    & $\checkmark$ & $\times$ & $\times$
    & \textbf{98.2} & 97.0 & \textbf{93.9} & 82.8 & 88.4 \\

    $\mathcal{M}_2$
    & $\checkmark$ & $\checkmark$ & $\times$
    & 97.9 & 97.4 & 93.8 & 84.0 & 88.9 \\

    $\mathcal{M}_3$
    & $\checkmark$ & $\checkmark$ & $\checkmark$
    & 97.9 & \textbf{98.6} & 93.8 & \textbf{85.4} & \textbf{90.1} \\

    \midrule

    $\mathcal{M}_{\mathrm{static}}$
    & \multicolumn{3}{c}{$\mathcal{M}_0$ + Static Cond.}
    & 97.9 & 96.7 & 93.6 & 82.3 & 87.8 \\

    \bottomrule
    \end{tabular}
}
}

\newcommand{\selectorablationtable}{%
\centering
\captionof{table}{
\textbf{\arranker{} ablation on NAVSIM v1 \texttt{navtest}.}
}
\label{tab:selector_ablation}
\setlength{\tabcolsep}{3.5pt}
\renewcommand{\arraystretch}{1.15}

\resizebox{\linewidth}{!}{
\begin{tabular}{l lll >{\columncolor[HTML]{E6E6E6}}l}
    \toprule
    \textbf{Selection}
    & \textbf{Gate} $\uparrow$
    & \textbf{TTC} $\uparrow$
    & \textbf{EP} $\uparrow$
    & \textbf{PDMS} $\uparrow$ \\
    \midrule

    \imranker
    & 94.78
    & 93.47
    & 82.44
    & 87.87 \\

    \arranker
    & \textbf{95.43}\pos{0.65}
    & \textbf{93.83}\pos{0.36}
    & \textbf{84.01}\pos{1.57}
    & \textbf{88.94}\pos{1.07} \\

    \bottomrule
\end{tabular}
}
}

\navsimresulttables

\section{EXPERIMENTS}
\label{sec:experiments}

\subsection{Experimental Setup}
\paragraph{Dataset.}
NAVSIM is an open-source benchmark for end-to-end planning built on the large-scale nuPlan dataset \cite{dauner2024navsim,nuplan}. It provides data from eight cameras and point clouds fused from five LiDAR sensors. We follow the official split, using the 1,192-scene \texttt{navtrain} set for training and the 136-scene \texttt{navtest} set for evaluation under NAVSIM v1 and v2.

\paragraph{NAVSIM metrics.}
NAVSIM v1 adopts a non-reactive simulation protocol. Under this protocol, a vehicle controller executes the planned trajectory, which is evaluated against the logged future using the PDM Score (PDMS). PDMS combines two multiplicative metrics, no-at-fault collision (NC) and drivable-area compliance (DAC), with three weighted metrics: time to collision (TTC), ride comfort (Comf.), and ego progress (EP). We report all component metrics and the overall PDMS.

NAVSIM v2 extends the evaluation with the Extended PDM Score (EPDMS), adding driving-direction compliance (DDC), traffic-light compliance (TLC), lane keeping (LK), historical comfort (HC), and extended comfort (EC) to NC, DAC, TTC, and EP. We report each component and the corresponding composite score.

\paragraph{Implementation details.}
\method{} takes multi-view RGB images and ego-state measurements as input and uses ResNet-34 as its visual backbone. The framework predicts $T=8$ future states at 2 Hz over a four-second horizon. The multimodal action source is a Gaussian mixture with $K=8$ components obtained by clustering the training set. We set the hidden dimension to $D=128$, use clearance margins of $\{0,0.5,1.0,2.0\}\,\mathrm m$, and set the field encoder's EMA decay to 0.999.

The model is trained end to end for 100 epochs with AdamW and an initial learning rate of $2\times10^{-4}$. The learning rate is linearly warmed up for the first three epochs and then decayed with a cosine schedule. Training uses eight NVIDIA H20 GPUs with a batch size of four per GPU. Pareto-ReinFlow is initialized from the complete supervised model and fine-tuned for 10 epochs, with learning rates of $10^{-5}$ and $3\times10^{-4}$ for the action head and exploration scale, respectively. All controlled experiments use the same visual backbone, multimodal source, and training data.

\subsection{Main Results}
Tables~\ref{tab:navsim_v1} and~\ref{tab:navsim_v2} compare \method{} with representative end-to-end driving methods. On NAVSIM v1, \method{} reaches 90.1 PDMS using only camera input, outperforming MeanFuser, one of the strongest prior MeanFlow method, by 1.1 points. Its leading DAC of 98.6 and EP of 85.4 show that the gain combines road compliance with forward progress. This result matches the role of the proposed representation: the configuration-space field explicitly models where the oriented ego footprint can move, while joint transport makes that evolving geometric constraint part of trajectory generation.

Without additional adaptation to the extended v2 metrics, \method{} still achieves the highest EPDMS of 89.7. It also attains the best DAC of 98.7 and ties for the best EP at 88.9, preserving the same strengths under the broader pseudo-closed-loop protocol. Overall, explicitly modeling future configuration feasibility, coupling it dynamically with trajectory transport, and applying safety-first Pareto optimization enable \method{} to maintain high progress and road compliance within a vision-only, one-step framework.

\subsection{Ablation Studies}
\paragraph{Ablation of key components.}
Table~\ref{tab:feasibleflow_ablation} progressively adds the core components of \method. Joint, ARR, and PT denote Asymmetric Joint MeanFlow, the \arranker{} (ARR), and Pareto-ReinFlow. The baseline $\mathcal M_0$ performs imitation learning with the GMM action source and one-step MeanFlow. Starting from this baseline, $\mathcal M_1$ adds joint transport, $\mathcal M_2$ adds the \arranker{}, and $\mathcal M_3$ adds Pareto-ReinFlow. The control $\mathcal M_{\mathrm{static}}$ retains feasibility reconstruction and conditions trajectory generation on the predicted field endpoint, but removes dynamic field--trajectory coupling. The baseline, $\mathcal M_1$, and $\mathcal M_{\mathrm{static}}$ share the same \imranker{}.

The static control remains at the baseline PDMS of 87.8, despite receiving both feasibility supervision and the predicted field endpoint. Merely adding a future representation is therefore insufficient: the representation must participate in trajectory formation. Joint transport raises PDMS to 88.4 and improves NC, DAC, TTC, and EP simultaneously over $\mathcal M_{\mathrm{static}}$. Because the two variants share the field representation, this broad improvement isolates the synchronous transport path and its cross-JVP term, which inject the direction of feasibility evolution into the trajectory target instead of exposing the generator only to a fixed endpoint.

The transition from $\mathcal M_2$ to $\mathcal M_3$ isolates generator post-training because the ranker remains frozen. Pareto-ReinFlow raises PDMS from 88.9 to 90.1, with DAC increasing from 97.4 to 98.6 and EP from 84.0 to 85.4 while preserving NC and TTC. The gain cannot arise from reordering the same slate; reference-relative Pareto feedback reshapes the action distribution toward candidates that jointly improve geometric compliance and progress.

\paragraph{\arranker.}
Table~\ref{tab:selector_ablation} compares the \imranker{} with the complete \arranker{} (ARR) on the same candidate set, isolating selection from generation. The complete ranker improves the joint safety gate $\mathrm{NC}\times\mathrm{DAC}$, TTC, and EP simultaneously, raising PDMS by 1.07 points. Because the candidate distribution is fixed, the result demonstrates that the gain comes from using multimodal candidates more effectively rather than generating an easier slate. The imitation anchor defines a scene-relative safety reference, while the outcome branch exploits the remaining diversity to select greater progress without predicted safety degradation. Together, the two ablations reveal a clear division of labor: the \arranker{} improves which candidate is chosen, whereas Pareto-ReinFlow improves which candidates are generated.

\begin{table}[t]
    \centering
    \componentablationtable

    \vspace{1.0em}

    \selectorablationtable
\end{table}

\subsection{Qualitative Results}
\begin{figure}[t]
    \centering
    \includegraphics[width=\columnwidth]{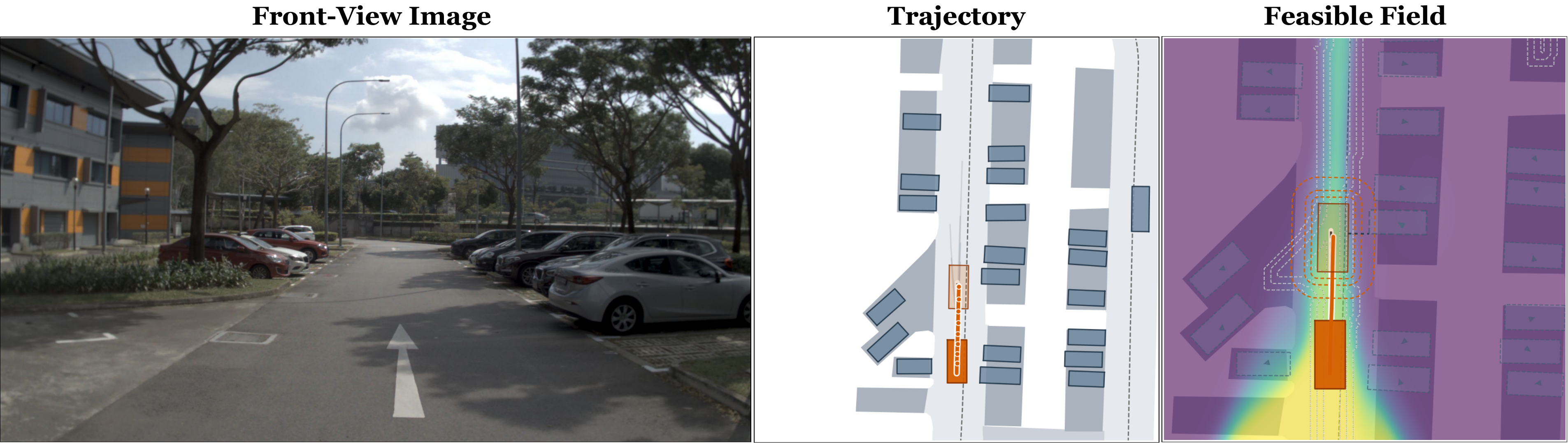}
    \captionof{figure}{\textbf{Visualization of configuration-space feasibility.} From left to right: the front-camera image, multimodal trajectory candidates with the selected plan, and the predicted feasibility field. Brighter regions in the heatmap indicate higher predicted feasibility. The field view overlays the exact configuration-space boundary, selected and ground-truth trajectories.}
    \label{fig:feasible_visualization}

    \vspace{0.8em}

    \includegraphics[width=\columnwidth]{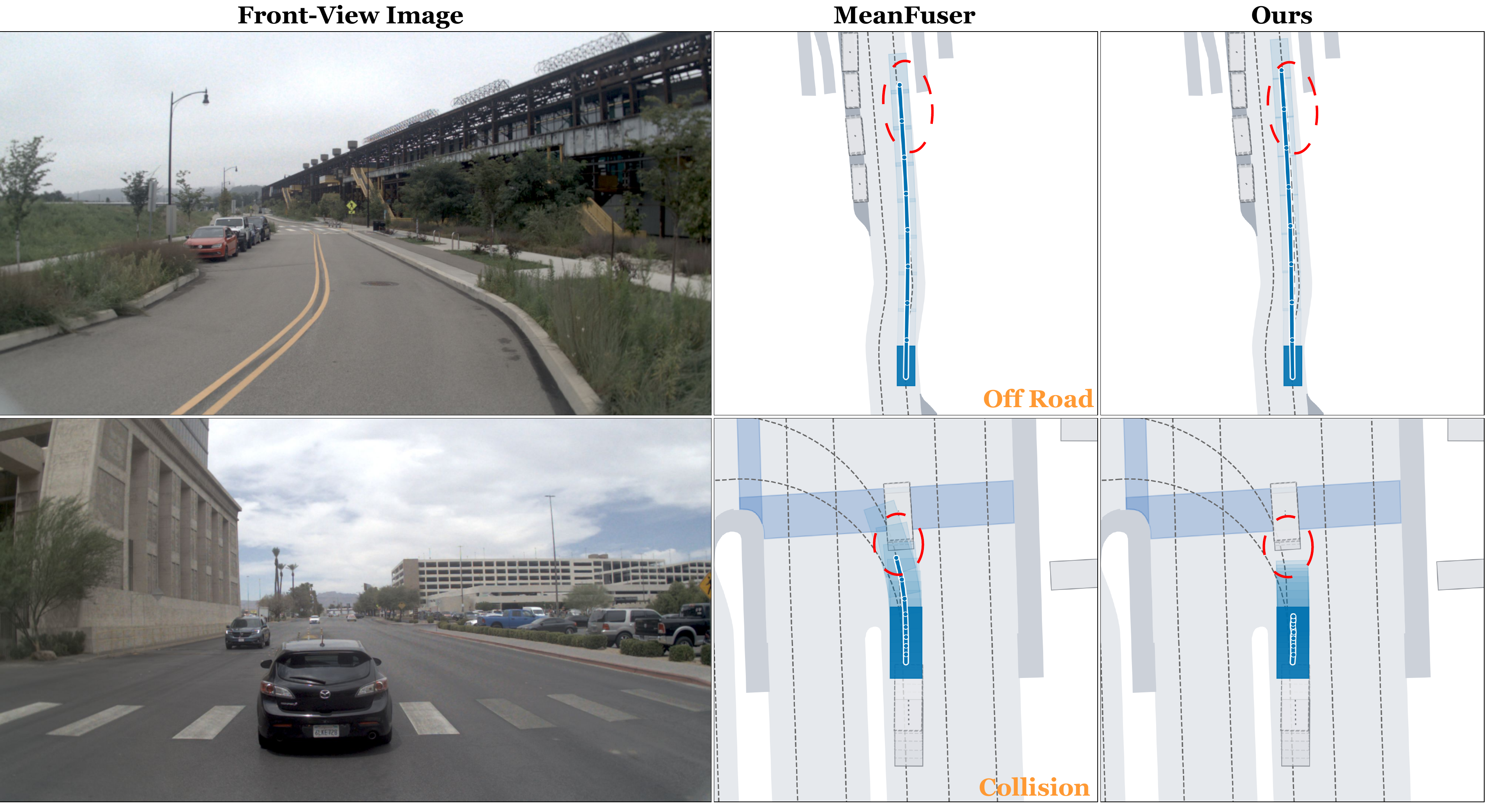}
    \captionof{figure}{\textbf{Qualitative comparison with MeanFuser~\cite{wang2026meanfuser}.} The top and bottom rows show off-road and collision cases, respectively. From left to right, each row presents the front-camera image, the MeanFuser result, and the \method{} result. Red circles mark the critical regions.}
    \label{fig:qualitative_comparison}
\end{figure}

In Figure~\ref{fig:feasible_visualization}, the bright, high-feasibility corridor follows the drivable passage and contracts near the interacting vehicle. The selected trajectory remains within this corridor and avoids the low-feasibility region around the vehicle, showing how the generated plan follows the spatial variation of predicted feasibility.

Figure~\ref{fig:qualitative_comparison} presents representative off-road and collision cases. In the upper example, MeanFuser crosses the drivable boundary as the road bends, whereas \method{} follows the curvature and remains within the lane. In the lower example, the baseline trajectory enters the leading vehicle's occupied configuration, while \method{} preserves longitudinal clearance. The two cases illustrate the complementary constraints encoded by the field: drivable-space geometry and clearance from other vehicles.

\section{CONCLUSION}
\label{sec:conclusion}
We propose \method, a one-step end-to-end driving framework that couples future configuration-space feasibility with multimodal trajectory generation. It comprises a graded configuration-space representation of future geometric constraints, an Asymmetric Joint MeanFlow that injects synchronous field evolution into trajectory transport, and a safety-first strategy combining the \arranker{} (ARR) with Pareto-ReinFlow. On NAVSIM, \method{} achieves leading vision-only planning performance with one-step inference, particularly in road compliance and ego progress. Ablations attribute these gains to dynamic feasibility--trajectory coupling and show complementary benefits from safety-first candidate selection and distribution optimization.

\bibliographystyle{IEEEtran}
\bibliography{conference}

@inproceedings{hu2022stp3,
  author    = {Hu, Shengchao and Chen, Li and Wu, Penghao and Li, Hongyang and Yan, Junchi and Tao, Dacheng},
  title     = {{ST-P3}: End-to-End Vision-Based Autonomous Driving via Spatial-Temporal Feature Learning},
  booktitle = {European Conference on Computer Vision},
  pages     = {533--549},
  year      = {2022},
  doi       = {10.1007/978-3-031-19839-7\_31}
}

@inproceedings{hu2023uniad,
  author    = {Hu, Yihan and Yang, Jiazhi and Chen, Li and Li, Keyu and Sima, Chonghao and Zhu, Xizhou and Chai, Siqi and Du, Senyao and Lin, Tianwei and Wang, Wenhai and Lu, Lewei and Jia, Xiaosong and Liu, Qiang and Dai, Jifeng and Qiao, Yu and Li, Hongyang},
  title     = {Planning-Oriented Autonomous Driving},
  booktitle = {Proceedings of the IEEE/CVF Conference on Computer Vision and Pattern Recognition},
  pages     = {17853--17862},
  year      = {2023}
}

@inproceedings{jiang2023vad,
  author    = {Jiang, Bo and Chen, Shaoyu and Xu, Qing and Liao, Bencheng and Chen, Jiajie and Zhou, Helong and Zhang, Qian and Liu, Wenyu and Huang, Chang and Wang, Xinggang},
  title     = {{VAD}: Vectorized Scene Representation for Efficient Autonomous Driving},
  booktitle = {Proceedings of the IEEE/CVF International Conference on Computer Vision},
  pages     = {8340--8350},
  year      = {2023}
}

@inproceedings{wang2023interpretable,
  author    = {Wang, Bikun and Wang, Zhipeng and Zhu, Chenhao and Zhang, Zhiqiang and Wang, Zhichen and Lin, Penghong and Liu, Jingchu and Zhang, Qian},
  title     = {Interpretable Motion Planner for Urban Driving via Hierarchical Imitation Learning},
  booktitle = {2023 IEEE/RSJ International Conference on Intelligent Robots and Systems},
  pages     = {1691--1696},
  year      = {2023},
  doi       = {10.1109/IROS55552.2023.10342448}
}

@inproceedings{zheng2024genad,
  author    = {Zheng, Wenzhao and Song, Ruiqi and Guo, Xianda and Zhang, Chenming and Chen, Long},
  title     = {{GenAD}: Generative End-to-End Autonomous Driving},
  booktitle = {European Conference on Computer Vision},
  pages     = {87--104},
  year      = {2024},
  doi       = {10.1007/978-3-031-73650-6\_6}
}

@inproceedings{zheng2024occworld,
  author    = {Zheng, Wenzhao and Chen, Weiliang and Huang, Yuanhui and Zhang, Borui and Duan, Yueqi and Lu, Jiwen},
  title     = {{OccWorld}: Learning a 3D Occupancy World Model for Autonomous Driving},
  booktitle = {European Conference on Computer Vision},
  pages     = {55--72},
  year      = {2024},
  doi       = {10.1007/978-3-031-72624-8\_4}
}

@inproceedings{yang2024diffusiones,
  author    = {Yang, Brian and Su, Huangyuan and Gkanatsios, Nikolaos and Ke, Tsung-Wei and Jain, Ayush and Schneider, Jeff and Fragkiadaki, Katerina},
  title     = {Diffusion-{ES}: Gradient-Free Planning with Diffusion for Autonomous and Instruction-Guided Driving},
  booktitle = {Proceedings of the IEEE/CVF Conference on Computer Vision and Pattern Recognition},
  pages     = {15342--15353},
  year      = {2024}
}

@inproceedings{liao2025diffusiondrive,
  author    = {Liao, Bencheng and Chen, Shaoyu and Yin, Haoran and Jiang, Bo and Wang, Cheng and Yan, Sixu and Zhang, Xinbang and Li, Xiangyu and Zhang, Ying and Zhang, Qian and Wang, Xinggang},
  title     = {DiffusionDrive: Truncated Diffusion Model for End-to-End Autonomous Driving},
  booktitle = {Proceedings of the IEEE/CVF Conference on Computer Vision and Pattern Recognition},
  pages     = {12037--12047},
  year      = {2025}
}

@inproceedings{xing2025goalflow,
  author    = {Xing, Zebin and Zhang, Xingyu and Hu, Yang and Jiang, Bo and He, Tong and Zhang, Qian and Long, Xiaoxiao and Yin, Wei},
  title     = {GoalFlow: Goal-Driven Flow Matching for Multimodal Trajectories Generation in End-to-End Autonomous Driving},
  booktitle = {Proceedings of the IEEE/CVF Conference on Computer Vision and Pattern Recognition},
  pages     = {1602--1611},
  year      = {2025}
}

@article{zou2025diffusiondrivev2,
  author  = {Zou, Jialv and Chen, Shaoyu and Liao, Bencheng and Zheng, Zhiyu and Song, Yuehao and Zhang, Lefei and Zhang, Qian and Liu, Wenyu and Wang, Xinggang},
  title   = {DiffusionDriveV2: Reinforcement Learning-Constrained Truncated Diffusion Modeling in End-to-End Autonomous Driving},
  journal = {arXiv preprint arXiv:2512.07745},
  year    = {2025},
  doi     = {10.48550/arXiv.2512.07745}
}

@inproceedings{wang2026meanfuser,
  author    = {Wang, Junli and Zheng, Yinan and Liu, Xueyi and Xing, Zebin and Li, Pengfei and Ma, Kun and Ye, Hangjun and Chen, Guang and Li, Guang and Chen, Long and Xia, Zhongpu and Zhang, Qichao},
  title     = {MeanFuser: Fast One-Step Multi-Modal Trajectory Generation and Adaptive Reconstruction via MeanFlow for End-to-End Autonomous Driving},
  booktitle = {Proceedings of the IEEE/CVF Conference on Computer Vision and Pattern Recognition},
  pages     = {17884--17893},
  year      = {2026}
}

@inproceedings{lipman2023flow,
  author    = {Lipman, Yaron and Chen, Ricky T. Q. and Ben-Hamu, Heli and Nickel, Maximilian and Le, Matt},
  title     = {Flow Matching for Generative Modeling},
  booktitle = {International Conference on Learning Representations},
  year      = {2023}
}

@inproceedings{geng2025meanflow,
  author    = {Geng, Zhengyang and Deng, Mingyang and Bai, Xingjian and Kolter, J. Zico and He, Kaiming},
  title     = {Mean Flows for One-Step Generative Modeling},
  booktitle = {Advances in Neural Information Processing Systems},
  volume    = {38},
  year      = {2025},
  doi       = {10.52202/085713-2534}
}

@article{lozanoperez1983,
  author  = {Lozano-P{\'e}rez, Tom{\'a}s},
  title   = {Spatial Planning: A Configuration Space Approach},
  journal = {IEEE Transactions on Computers},
  volume  = {C-32},
  number  = {2},
  pages   = {108--120},
  year    = {1983},
  doi     = {10.1109/TC.1983.1676196}
}

@inproceedings{ratliff2009chomp,
  author    = {Ratliff, Nathan and Zucker, Matt and Bagnell, J. Andrew and Srinivasa, Siddhartha},
  title     = {{CHOMP}: Gradient Optimization Techniques for Efficient Motion Planning},
  booktitle = {2009 IEEE International Conference on Robotics and Automation},
  pages     = {489--494},
  year      = {2009},
  doi       = {10.1109/ROBOT.2009.5152817}
}

@inproceedings{dauner2024navsim,
  author    = {Dauner, Daniel and Hallgarten, Marcel and Li, Tianyu and Weng, Xinshuo and Huang, Zhiyu and Yang, Zetong and Li, Hongyang and Gilitschenski, Igor and Ivanovic, Boris and Pavone, Marco and Geiger, Andreas and Chitta, Kashyap},
  title     = {{NAVSIM}: Data-Driven Non-Reactive Autonomous Vehicle Simulation and Benchmarking},
  booktitle = {Advances in Neural Information Processing Systems},
  volume    = {37},
  pages     = {28706--28719},
  year      = {2024},
  doi       = {10.52202/079017-0902}
}

@article{chitta2022transfuser,
  author  = {Chitta, Kashyap and Prakash, Aditya and Jaeger, Bernhard and Yu, Zehao and Renz, Katrin and Geiger, Andreas},
  title   = {{TransFuser}: Imitation with Transformer-Based Sensor Fusion for Autonomous Driving},
  journal = {IEEE Transactions on Pattern Analysis and Machine Intelligence},
  volume  = {45},
  number  = {11},
  pages   = {12878--12895},
  year    = {2023},
  doi     = {10.1109/TPAMI.2022.3200245}
}

@inproceedings{weng2024paradrive,
  author    = {Weng, Xinshuo and Ivanovic, Boris and Wang, Yan and Wang, Yue and Pavone, Marco},
  title     = {{PARA-Drive}: Parallelized Architecture for Real-Time Autonomous Driving},
  booktitle = {Proceedings of the IEEE/CVF Conference on Computer Vision and Pattern Recognition},
  pages     = {15449--15458},
  year      = {2024},
  doi       = {10.1109/CVPR52733.2024.01463}
}

@inproceedings{sun2024sparsedrive,
  author    = {Sun, Wenchao and Lin, Xuewu and Shi, Yining and Zhang, Chuang and Wu, Haoran and Zheng, Sifa},
  title     = {{SparseDrive}: End-to-End Autonomous Driving via Sparse Scene Representation},
  booktitle = {2025 IEEE International Conference on Robotics and Automation},
  pages     = {8795--8801},
  year      = {2025},
  doi       = {10.1109/ICRA55743.2025.11128800}
}

@inproceedings{yang2025driveoccworld,
  author    = {Yang, Yu and Mei, Jianbiao and Ma, Yukai and Du, Siliang and Chen, Wenqing and Qian, Yijie and Feng, Yuxiang and Liu, Yong},
  title     = {Driving in the Occupancy World: Vision-Centric 4D Occupancy Forecasting and Planning via World Models for Autonomous Driving},
  booktitle = {Proceedings of the AAAI Conference on Artificial Intelligence},
  volume    = {39},
  number    = {9},
  pages     = {9327--9335},
  year      = {2025},
  doi       = {10.1609/aaai.v39i9.33010}
}

@inproceedings{zheng2025world4drive,
  author    = {Zheng, Yupeng and Yang, Pengxuan and Xing, Zebin and Zhang, Qichao and Zheng, Yuhang and Gao, Yinfeng and Li, Pengfei and Zhang, Teng and Xia, Zhongpu and Jia, Peng and Lang, XianPeng and Zhao, Dongbin},
  title     = {{World4Drive}: End-to-End Autonomous Driving via Intention-Aware Physical Latent World Model},
  booktitle = {Proceedings of the IEEE/CVF International Conference on Computer Vision},
  pages     = {28632--28642},
  year      = {2025}
}

@inproceedings{li2025wote,
  author    = {Li, Yingyan and Wang, Yuqi and Liu, Yang and He, Jiawei and Fan, Lue and Zhang, Zhaoxiang},
  title     = {End-to-End Driving with Online Trajectory Evaluation via {BEV} World Model},
  booktitle = {Proceedings of the IEEE/CVF International Conference on Computer Vision},
  pages     = {27137--27146},
  year      = {2025}
}

@inproceedings{zheng2025diffusionplanner,
  author    = {Zheng, Yinan and Liang, Ruiming and Zheng, Kexin and Zheng, Jinliang and Mao, Liyuan and Li, Jianxiong and Gu, Weihao and Ai, Rui and Li, Shengbo Eben and Zhan, Xianyuan and Liu, Jingjing},
  title     = {Diffusion-Based Planning for Autonomous Driving with Flexible Guidance},
  booktitle = {International Conference on Learning Representations},
  year      = {2025}
}

@inproceedings{liu2025guideflow,
  author    = {Liu, Lin and Jia, Caiyan and Yu, Guanyi and Song, Ziying and Li, Junqiao and Jia, Feiyang and Wu, Peiliang and Hao, Xiaoshuai and Luo, Yadan},
  title     = {{GuideFlow}: Constraint-Guided Flow Matching for Planning in End-to-End Autonomous Driving},
  booktitle = {Proceedings of the IEEE/CVF Conference on Computer Vision and Pattern Recognition},
  pages     = {3719--3728},
  year      = {2026}
}

@inproceedings{xu2026wamflow,
  author    = {Xu, Yifang and Cui, Jiahao and Zhu, Zhihao and Shang, Hanlin and Luan, Shan and Xu, Mingwang and Cai, Feipeng and Zhang, Neng and Li, Yaoyi and Cai, Jia and Zhu, Siyu},
  title     = {{WAM-Flow}: Parallel Coarse-to-Fine Motion Planning via Discrete Flow Matching for Autonomous Driving},
  booktitle = {Proceedings of the IEEE/CVF Conference on Computer Vision and Pattern Recognition},
  pages     = {24918--24928},
  year      = {2026}
}

@inproceedings{zhang2025reinflow,
  author    = {Zhang, Tonghe and Yu, Chao and Su, Sichang and Wang, Yu},
  title     = {{ReinFlow}: Fine-Tuning Flow Matching Policy with Online Reinforcement Learning},
  booktitle = {Advances in Neural Information Processing Systems},
  volume    = {38},
  year      = {2025},
  doi       = {10.52202/085713-3547}
}

@inproceedings{shang2025drivedpo,
  author    = {Shang, Shuyao and Chen, Yuntao and Wang, Yuqi and Li, Yingyan and Zhang, Zhaoxiang},
  title     = {{DriveDPO}: Policy Learning via Safety {DPO} for End-to-End Autonomous Driving},
  booktitle = {Advances in Neural Information Processing Systems},
  volume    = {38},
  year      = {2025},
  doi       = {10.52202/085713-2724}
}

@inproceedings{zhang2025seerdrive,
  author    = {Zhang, Bozhou and Song, Nan and Li, Jingyu and Zhu, Xiatian and Deng, Jiankang and Zhang, Li},
  title     = {Future-Aware End-to-End Driving: Bidirectional Modeling of Trajectory Planning and Scene Evolution},
  booktitle = {Advances in Neural Information Processing Systems},
  volume    = {38},
  year      = {2025},
  doi       = {10.52202/085713-0345}
}

@inproceedings{fu2026prodrive,
  author    = {Fu, Chuyao and Gan, Shengzhe and Ouyang, Zhuoli and Rui, Yuhan and Chi, Xiaowei and Han, Sirui and Wang, Jiankun and Zhang, Hong},
  title     = {{ProDrive}: Proactive Planning for Autonomous Driving via Ego-Environment Co-Evolution},
  booktitle = {Proceedings of the IEEE/CVF Conference on Computer Vision and Pattern Recognition Workshops},
  pages     = {4436--4445},
  year      = {2026}
}

@article{li2024hydramdp,
  author  = {Li, Zhenxin and Li, Kailin and Wang, Shihao and Lan, Shiyi and Yu, Zhiding and Ji, Yishen and Li, Zhiqi and Zhu, Ziyue and Kautz, Jan and Wu, Zuxuan and Jiang, Yu-Gang and Alvarez, Jose M.},
  title   = {{Hydra-MDP}: End-to-End Multimodal Planning with Multi-Target Hydra-Distillation},
  journal = {arXiv preprint arXiv:2406.06978},
  year    = {2024},
  doi     = {10.48550/arXiv.2406.06978}
}

@article{li2025hydramdpp,
  author  = {Li, Kailin and Li, Zhenxin and Lan, Shiyi and Xie, Yuan and Zhang, Zhizhong and Liu, Jiayi and Wu, Zuxuan and Yu, Zhiding and Alvarez, Jose M.},
  title   = {{Hydra-MDP++}: Advancing End-to-End Driving via Expert-Guided Hydra-Distillation},
  journal = {arXiv preprint arXiv:2503.12820},
  year    = {2025},
  doi     = {10.48550/arXiv.2503.12820}
}

@inproceedings{zhang2025epona,
  author    = {Zhang, Kaiwen and Tang, Zhenyu and Hu, Xiaotao and Pan, Xingang and Guo, Xiaoyang and Liu, Yuan and Huang, Jingwei and Yuan, Li and Zhang, Qian and Long, Xiao-Xiao and Cao, Xun and Yin, Wei},
  title     = {{Epona}: Autoregressive Diffusion World Model for Autonomous Driving},
  booktitle = {Proceedings of the IEEE/CVF International Conference on Computer Vision},
  pages     = {27220--27230},
  year      = {2025},
  doi       = {10.1109/ICCV51701.2025.02527}
}

@inproceedings{yao2026drivesuprim,
  author    = {Yao, Wenhao and Li, Zhenxin and Lan, Shiyi and Wang, Zi and Sun, Xinglong and Alvarez, Jose M. and Wu, Zuxuan},
  title     = {{DriveSuprim}: Towards Precise Trajectory Selection for End-to-End Planning},
  booktitle = {Proceedings of the AAAI Conference on Artificial Intelligence},
  volume    = {40},
  number    = {14},
  pages     = {11910--11918},
  year      = {2026},
  doi       = {10.1609/aaai.v40i14.38178}
}

@article{wang2026drivejepa,
  author  = {Wang, Linhan and Yang, Zichong and Bai, Chen and Zhang, Guoxiang and Liu, Xiaotong and Zheng, Xiaoyin and Long, Xiao-Xiao and Lu, Chang-Tien and Lu, Cheng},
  title   = {{Drive-JEPA}: Video {JEPA} Meets Multimodal Trajectory Distillation for End-to-End Driving},
  journal = {arXiv preprint arXiv:2601.22032},
  year    = {2026},
  doi     = {10.48550/arXiv.2601.22032}
}

@inproceedings{nuplan,
  author    = {Caesar, Holger and Kabzan, Juraj and Tan, Kok Seang and Fong, Whye Kit and Wolff, Eric M. and Lang, Alex H. and Fletcher, Luke and Beijbom, Oscar and Omari, Sammy},
  title     = {{nuPlan}: A Closed-Loop ML-Based Planning Benchmark for Autonomous Vehicles},
  booktitle = {CVPR Workshop on Autonomous Driving: Perception, Prediction and Planning},
  year      = {2021}
}

\end{document}